\documentclass[11pt]{article}

\usepackage{acl}

\usepackage{times}
\usepackage{latexsym}

\usepackage{amsmath}
\usepackage{amssymb}   

\usepackage[T1]{fontenc}
\usepackage[utf8]{inputenc}
\usepackage{microtype}
\usepackage{inconsolata}
\usepackage{rotating}
\usepackage{graphicx}
\usepackage{booktabs}
\usepackage{array}
\usepackage{tabularray}
\usepackage{pdflscape}
\usepackage{longtable}
\usepackage[table]{xcolor}
\usepackage[most]{tcolorbox}
\usepackage{float}
\usepackage{subcaption}

\newtcolorbox{promptbox}[2][]{%
  breakable, enhanced, colback=white, colframe=black!55,
  coltitle=white, colbacktitle=black!60, fonttitle=\bfseries\small,
  title=#2, boxrule=0.6pt, arc=3pt, left=5pt, right=5pt, top=4pt, bottom=4pt,
  before skip=6pt, after skip=6pt, #1}

\title{QuranicMMLU: A Cognitively-Aware Benchmark for Evaluating Generative AI Solutions on Quranic Linguistic Knowledge}

\author{
  Rawan El Ghali\textsuperscript{1,2}\footnotemark[1] \quad
  Umm Kulsoom\textsuperscript{1,3}\footnotemark[1] \quad
  Anas Madkoor\textsuperscript{1} \quad
  Dima Faris Alsaudi\textsuperscript{1,4} \\
  \bfseries Roaa Abdelmagid\textsuperscript{1,6} \quad
  Roaa Ibrahim\textsuperscript{1,3} \quad
  Raghad Mousa\textsuperscript{1,2} \\
  \bfseries Hamza Aljaji\textsuperscript{1,4} \quad
  Abdullah Khanafer\textsuperscript{1,5} \quad
  Abdallah Alkanani\textsuperscript{1,4} \\
  \bfseries Salah Feras Alali\textsuperscript{1,4} \quad
  Rawan Khaled Mohamed\textsuperscript{1,4} \quad
  Ehsaneddin Asgari\textsuperscript{1} \\[0.25em]
  {\normalfont \textsuperscript{1}Qatar Computing Research Institute, Hamad Bin Khalifa University \quad
  \textsuperscript{2}Carnegie Mellon University Qatar} \\
  {\normalfont \textsuperscript{3}University of Doha for Science and Technology \quad
  \textsuperscript{4}Qatar University} \\
  {\normalfont \textsuperscript{5}American University of Beirut \quad
  \textsuperscript{6}University of Birmingham} \\
  {\normalfont\small \texttt{easgari@hbku.edu.qa}} \\[0.5em]
}

\begin{document}
\raggedbottom
\renewcommand{\thefootnote}{}
\maketitle
\renewcommand{\thefootnote}{\arabic{footnote}}
\setcounter{footnote}{0}

\begingroup
\renewcommand{\thefootnote}{\ensuremath{*}}
\footnotetext[1]{\;Equal contribution; joint first authors.}
\endgroup

\begin{abstract}
We introduce QuranicMMLU, a benchmark for evaluating generative AI on Quranic
Arabic across multiple dimensions of linguistic complexity. Existing Quranic 
benchmarks center on general question answering and semantic retrieval, without
probing specific linguistic competencies or stratifying by cognitive demand and
verse difficulty. We construct a five-pillar Quranic taxonomy spanning Phonology,
Morphology, Syntax, Semantics, and Pragmatics, with 31 leaves
covering phenomena from tajw\={\i}d and root-and-pattern morphology to occasions of revelation and inter-surah coherence. For each leaf we generate questions
stratified by Bloom's cognitive level and verse perplexity, then have LLM as a judge
to independently answer and score every item and route the annotations
to manual review. The resulting dataset comprises 980 human-reviewed questions,
each issued in both open-ended and multiple-choice form. We benchmark 12 systems
on these items and find that the Islamic-specialized model leads, yet every system
scores higher on multiple-choice accuracy (average 84\%) than open-ended answer quality (average 60\%): the two rankings agree closely (Kendall's $\tau=0.73$), but
multiple-choice scoring hides failures that surface only once answer choices are
removed. QuranicMMLU thus offers a rigorous, linguistically grounded framework for
evaluating Arabic NLP in the Quranic domain.
\end{abstract}

\section{Introduction}
The Quran is the primary religious and cultural text for nearly two
billion Muslims, and engaging with it meaningfully demands not
just reading but a genuine understanding of its language and content. As
large language models are increasingly consulted on Islamic topics,
including direct questions about the Quran
 \citep{abdelaal2026islamicmmlu}, the cost of
ungrounded answers in this domain is unusually high: hallucinated or
inaccurate responses carry real religious consequences
 \citep{mubarak2025islamiceval}. Reliable use of these systems
therefore depends on accurate, fine-grained evaluation of how well they
actually understand Quranic Arabic.

Existing benchmarks only partially meet this need. Islamic and Quranic
QA resources assess factual and reasoning ability
 \citep{abdelaal2026islamicmmlu}. Three gaps remain.
First, no benchmark targets the diversity of \emph{linguistic}
competencies, from phonology to pragmatics, that Quranic understanding
requires. Second, most datasets are multiple-choice only, a format that
can be gamed through answer-choice artifacts
 \citep{balepur2024artifacts} and that diverges from how users actually
query models. Third, the cognitive demand of questions is rarely
measured systematically.

We introduce QuranicMMLU to address these gaps. Our contributions are:
\begin{enumerate}
    \renewcommand{\labelenumi}{(\roman{enumi})}
    \item A five-pillar taxonomy of Quranic Arabic spanning Phonology,
           Morphology, Syntax, Semantics, and Pragmatics, with 31 leaves
           covering competencies from tajw\={\i}d to inter-surah coherence;
     \item A human-reviewed, retrieval-grounded dataset of 980 questions,
           each in both open-ended and multiple-choice form and stratified by
           Bloom's cognitive level \citep{anderson2001taxonomy} and verse `popularity of use', measured by perplexity;
    \item An evaluation of 12 systems in both open-ended and multiple-choice formats, showing that multiple-choice scoring overstates ability and hides failures that surface only in open-ended use.
\end{enumerate}

The leaderboard and relevant resources are available at
\url{https://huggingface.co/spaces/musiml-org/QuranicMMLU}.

\section{Related Work}
\subsection{Islamic and Quranic QA Benchmarks}
 Early Quranic QA frames the task as extractive answer-span retrieval
  \citep{MALHAS2022103068, malhas-etal-2022-quran, malhas-etal-2023-quran,
 alnefaie-etal-2023-haqa}, while recent benchmarks broaden to Islamic knowledge
 and reasoning through multiple-choice, retrieval-augmented, and shared-task
 settings  \citep{abdelaal2026islamicmmlu,
 bhatia2026ragagenticragfaithful, qias2025, alwajih-etal-2025-palm}, and extend to
 Islamic jurisprudence across the Sunni legal schools
  \citep{atif2025sacred, bahaj2025mizanqa}. These resources
 target factual and reasoning ability; we instead probe Quranic
 \emph{linguistic} competencies, from phonology to pragmatics, in both
 multiple-choice and open-ended formats.

 \subsection{Arabic Understanding Benchmarks}
 General Arabic benchmarks such as ArabicMMLU and AlGhafa measure broad
 competence \citep{koto2024arabicmmlu,
 almazrouei-etal-2023-alghafa}, but not the phonological, morphological,
 syntactic, and rhetorical phenomena specific to Quranic Arabic that we target.

 \subsection{Cognitive and Format-Aware Evaluation}
Bloom's taxonomy \cite{krathwohl2002revision} has become a foundational scaffold for stratifying cognitive demands in language model evaluation, moving assessment beyond aggregate accuracy toward diagnosing where competence breaks down across cognition layers. Recent work has demonstrated the applicability of cognitive assessment frameworks across visual understanding \citep{abootorabi2026almiyear} and relational knowledge in multilingual contexts \citep{liasi2026reasoning}. However, most benchmarks remain limited: they concentrate on lower-order cognitive tasks, rely predominantly on multiple-choice formats vulnerable to answer-choice artifacts \citep{balepur2024artifacts}, and lack systematic frameworks for comprehensive cognitive evaluation. To address these gaps, we introduce QuranicMMLU, a benchmark that combines multiple question formats with a multi-judge protocol to systematically evaluate knowledge across cognitive levels and verse-perplexity dimensions.





\section{Approach} 
\begin{figure}[t]
   \centering
   \includegraphics[width=0.85\columnwidth]{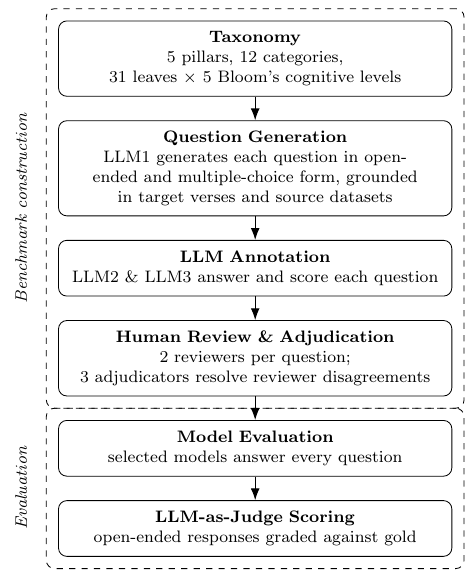}
   \vspace{-8pt}
   \caption{Overview of QuranicMMLU: taxonomy-driven question generation, LLM
   annotation and human review, then model evaluation with LLM-as-judge
   scoring. LLM1 is Claude (generator); LLM2 and LLM3 are GPT and Gemini
   (annotators).}
   \label{fig:approach-overview}
 \end{figure}






    
QuranicMMLU is built through a six-stage pipeline
(Figure~\ref{fig:approach-overview}): we define the taxonomy, generate questions per leaf, collect LLM annotations, conduct human review with conflict resolution, benchmark selected models, and score open-ended responses with an LLM judge. Each stage is detailed below.




\subsection{Taxonomies}
 \label{sec:taxonomy}
 \textbf{QuranicMMLU Taxonomy.}
 We built the taxonomy by studying the literature and key sources for each
 pillar. We drafted candidate categories and refined them with the help of AI
 deep-research tools, ensuring broad coverage of the target capability space by
 including both commonly studied phenomena and less frequently represented but
 conceptually important aspects. Related items were then grouped into the final
 taxonomy.

\begin{figure*}[p]
  \centering
  \includegraphics[width=\textwidth,height=0.92\textheight,keepaspectratio]{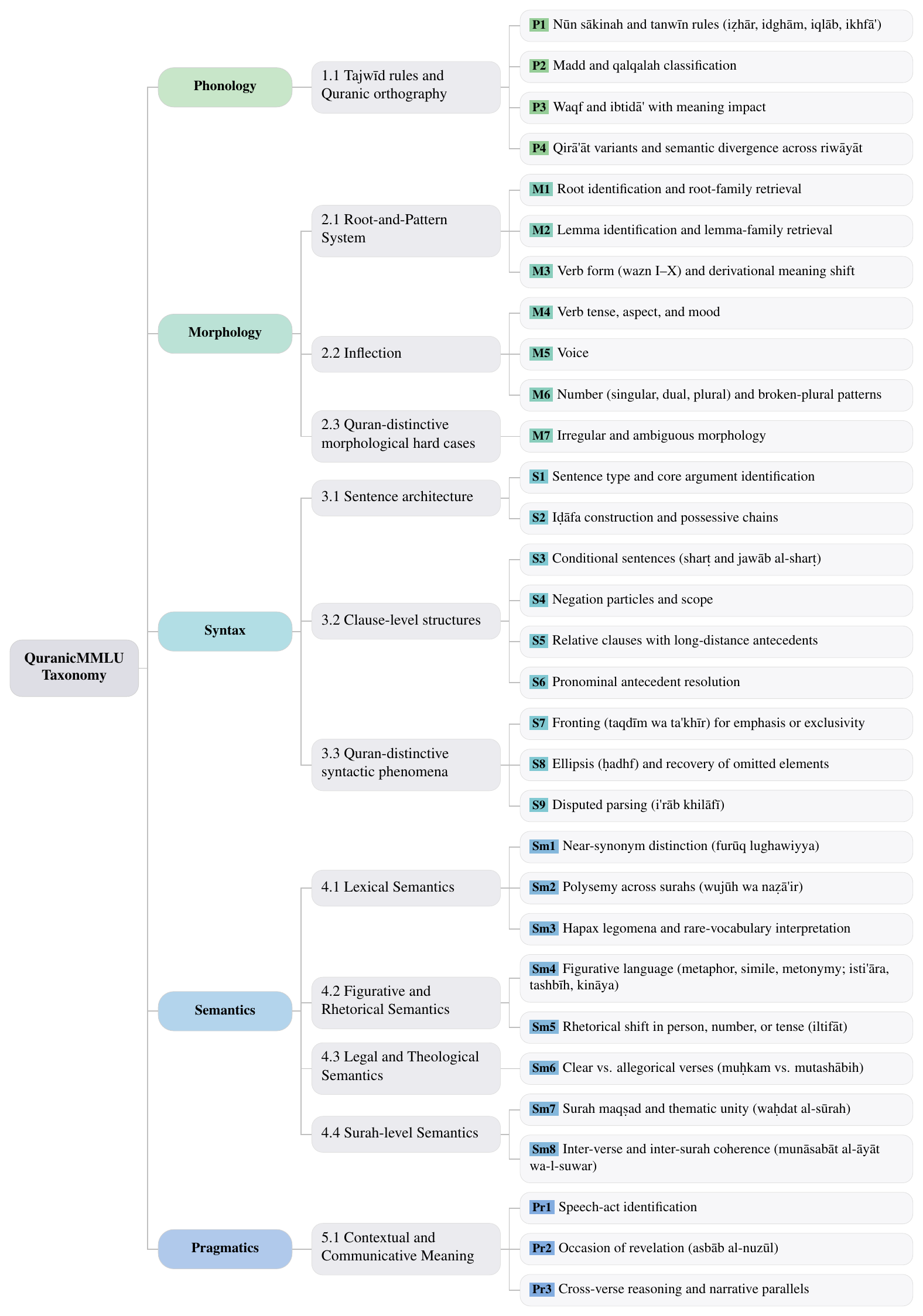}
  \caption{The QuranicMMLU taxonomy: 5 linguistic pillars,
  12 categories, and 31 leaves. Leaf codes (e.g., \texttt{P1},
  \texttt{Sm8}) index the specific phenomena evaluated by the benchmark.}
  \label{fig:taxonomy}
\end{figure*}

 We organize the QuranicMMLU Taxonomy as 5 pillars, 12 categories, and 31 leaves
 (Figure~\ref{fig:taxonomy}), each leaf a distinct phenomenon, so evaluation
 is a grid of competencies rather than one aggregate score. Unlike MMLU
  \citep{hendrycks2021mmlu} and ArabicMMLU  \citep{koto2024arabicmmlu}, which
 probe subject knowledge, we organize by linguistic ability
  \citep{ribeiro2020checklist}, spreading each pillar's leaves across a range of its sub-phenomena.

  \textit{Phonology} covers madd,
 qalqalah, waqf, and qir\=a\textquotesingle \=at, probing the
 sound--meaning mapping that Arabic's undiacritized script makes a hard
 disambiguation task  \citep{zitouni2006diac}. \textit{Morphology}
 spans roots, lemmas, wazn, tense--aspect--mood, voice, number, and
 clitics, all sources of the high ambiguity of the root-and-pattern system and grounded in the Quranic Arabic Corpus
 \citep{dukes-habash-2010-morphological}. \textit{Syntax} ranges
 from sentence type and i\d{d}\=afa to negation scope, fronting, ellipsis, and
 disputed i\textquotesingle r\=ab, reflecting the difficulty of Arabic parsing
  \citep{green-manning-2010-arabic} and grounded in the dependency treebank
  \citep{dukes-etal-2010-syntactic}. \textit{Semantics}, the widest
 pillar, covers lexical ambiguity (fur\=uq, wuj\=uh wa na\d{z}\=a\textquotesingle
 ir), figurative language, and surah-level meaning, all open problems in Arabic
 word-sense disambiguation  \citep{jarrar2023salma, raganato-etal-2017-word},
 figurative-language understanding \citep{tsvetkov-etal-2014-metaphor}, and
 discourse coherence  \citep{barzilay-lapata-2008-modeling}. \textit{Pragmatics}
 targets speech acts, asb\=ab al-nuz\=ul, and cross-verse reasoning, tasks
 needing communicative-function recognition \citep{stolcke-etal-2000-dialogue}
 and context beyond span retrieval  \citep{MALHAS2022103068, malhas-etal-2022-quran}.
Appendix~\ref{app:examples} (p.~\pageref{app:examples}) gives one
representative question per pillar with its verified Bloom level.

\textbf{Bloom's Cognitive Level Taxonomy.}
We stratify questions across five of the six levels of Bloom's
taxonomy, Remember, Understand, Apply, Analyze, and Evaluate, omitting
Create because generating new Quranic text raises clear ethical
concerns (Figure~\ref{fig:bloom} in Appendix~\ref{app:taxonomy-fig}). The operational test reviewers use to assign these
levels is given in Appendix~\ref{app:guidelines}
(p.~\pageref{app:guidelines}).

\subsection{Datasets}
\label{sec:datasets}
We ground questions and answers in authoritative sources:
the Quran.com API for \textbf{Phonology}; the Quranic Arabic Corpus for
\textbf{Morphology} \citep{dukes-habash-2010-morphological} and
\textbf{Syntax} \citep{dukes-etal-2010-syntactic}; Quranic tafs\={\i}r from
Tarteel\footnote{\url{https://huggingface.co/datasets/tarteel-ai/quran-tafsir}}
for \textbf{Semantics}; and an asb\=ab al-nuz\=ul
dataset\footnote{\url{https://github.com/mostafaahmed97/asbab-al-nuzul-dataset}}
based on \textit{Sahih Asbab al-Nuzul} (Ibrahim Muhammad al-Ali) for
\textbf{Pragmatics}.

\subsection{Question Generation and Annotation}
\label{sec:generation}
\textbf{Generation.} For each leaf, we use Claude Opus 4.7 to generate questions, reference
short answers, and multiple-choice options at each Bloom level. Generation
begins by selecting a verse or group of verses containing an instance of
the target leaf (for example, a verse with a simile or metaphor when
asking about figurative language). We build this pool by searching the
grounding datasets for verses that contain an instance of the target
phenomenon, at times using AI research tools to help find candidates, and
manually reviewing them to confirm the instance is present; generation then
draws a verse at random. We also balance questions across verse
perplexity (computed using the Gemma-3-12B model), a proxy for the model's familiarity with a verse, aiming for
comparable counts per leaf across the low, medium, and high
levels; this stratification is applied to single-verse questions rather than
whole-surah ones. The relevant verse is included in the prompt (except for
surah-level questions) to avoid misretrieval of Quranic text. Each
question is produced with a four-layer prompt stack
(Appendix~\ref{app:prompt-stack}, p.~\pageref{app:prompt-stack};
Figure~\ref{fig:prompt-stack}) in which
every leaf and Bloom level has a specialized prompt, and each item is
written to support both open-ended and multiple-choice answering. Of the
four options per question, one is correct, two are close distractors, and
one is clearly wrong. For both formats, the datasets (Section~\ref{sec:datasets}) ground every answer in an authoritative source
(a retrieval-augmented setup).

\textbf{LLM annotation.} The generated items are then annotated by two LLM judges, GPT-5.2 and
Gemini-2.5-Flash \citep{comanici2025gemini25pushingfrontier}. Each judge writes an open-ended answer, selects the
correct MCQ option, assigns a Bloom level, and scores the question out of
five on specificity, answerability, leaf relevance, and clarity
(Appendix~\ref{app:rubric}, p.~\pageref{app:rubric}). Both
judges are given full access to the grounding datasets (Section~\ref{sec:datasets}) and prompted to
base answers on them and cite their source otherwise, again making
annotation retrieval-augmented rather than memory-dependent. On the
multiple-choice answer, the three independently prompted models (the
generator and the two judges) show almost-perfect agreement (Fleiss's
$\kappa = 0.90$).

\textbf{Human review and adjudication.} Every annotation is then reviewed by
two humans per question to reduce individual bias. Reviewers are
undergraduate students, fluent Arabic speakers with a background in Islamic
knowledge; because they are not religious scholars, every decision is
grounded in corpus evidence and cited references rather than personal
judgment, following a written guide (Appendix~\ref{app:guidelines},
p.~\pageref{app:guidelines}) that ranks sources of truth: corpus evidence
first, then established scholarly consensus, then genuine dispute. Reviewers
may edit the question, correct its Bloom level, choose among the annotated
answers, add a new answer when none is correct, or delete items that are
irrelevant, incorrect, or controversial; because we prioritized correctness
and Islamic soundness, any item with a suspected hallucination, an
unresolved scholarly dispute, or a single flagged concern was discarded when
it could not be cleanly fixed. Counting any such edit as a revision, 370 of
the 1,086 candidates (34\%) were revised, most often a corrected Bloom level
(235 items), 106 (10\%) were discarded, and 610 were accepted as generated,
leaving the final 980. Conflicts were settled by majority verdict among
three adjudicators, with 133 items escalated to group discussion. The review
interface is shown in Appendix~\ref{app:review} (p.~\pageref{app:review}).

\begin{figure*}[t]
  \centering
  \includegraphics[width=0.95\textwidth]{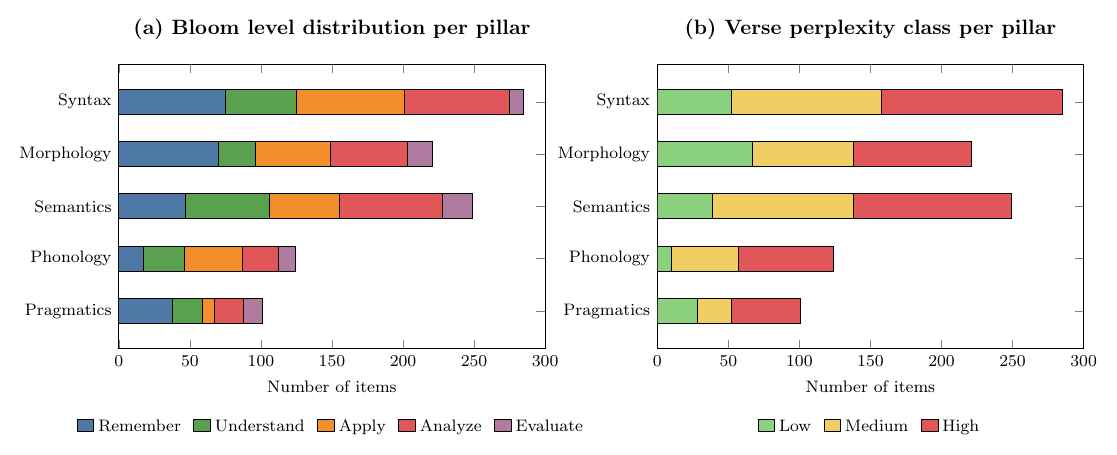}
  \caption{Dataset composition: (a) Bloom-level distribution per pillar and
  (b) verse-perplexity class per pillar. The overall Bloom-level
  distribution across all 980 items is given in
  Figure~\ref{fig:bloom-overall}.}
  \label{fig:dataset-overview}
\end{figure*}

\textbf{Dataset statistics.} Figure~\ref{fig:dataset-overview} reports the
distribution of the 980 questions, showing the Bloom-level and
verse-perplexity composition within each pillar. The dataset spans all five
pillars and all five Bloom levels, with Syntax, Semantics, and Morphology
the most represented pillars and a deliberate concentration of
higher-perplexity verses, so that evaluation is not dominated by verses
models have likely memorized.

\subsection{Benchmarking}
We benchmark the final dataset in both formats, MCQ and open-ended,
over the identical item pool, so results are directly comparable across
the two. We evaluate two classes of systems: (i) \textbf{open-source
models}, general-purpose open-weight LLMs including Gemma (Google)
\citep{gemmateam2025gemma3technicalreport,gemmateam2026gemma4technicalreport},
Qwen/Qwen-VL (Alibaba)
\citep{yang2025qwen3technicalreport,bai2025qwen3vltechnicalreport}, and the
Arabic-centric ALLaM \citep{bari2024allamlargelanguagemodels}; and (ii)
\textbf{Islamic models} specialized for Quranic content, namely
Fanar-Sadiq, the Islamic-oriented model from Fanar (QCRI)
\citep{fanarteam2025fanararabiccentricmultimodalgenerative,fanarteam2026fanar20arabicgenerative},
and Ansari \citep{kadous2026ansariretrievalgroundedislamicai}.

 \subsection{Metrics}
\begin{figure*}[t]
  \centering
  \includegraphics[width=\textwidth]{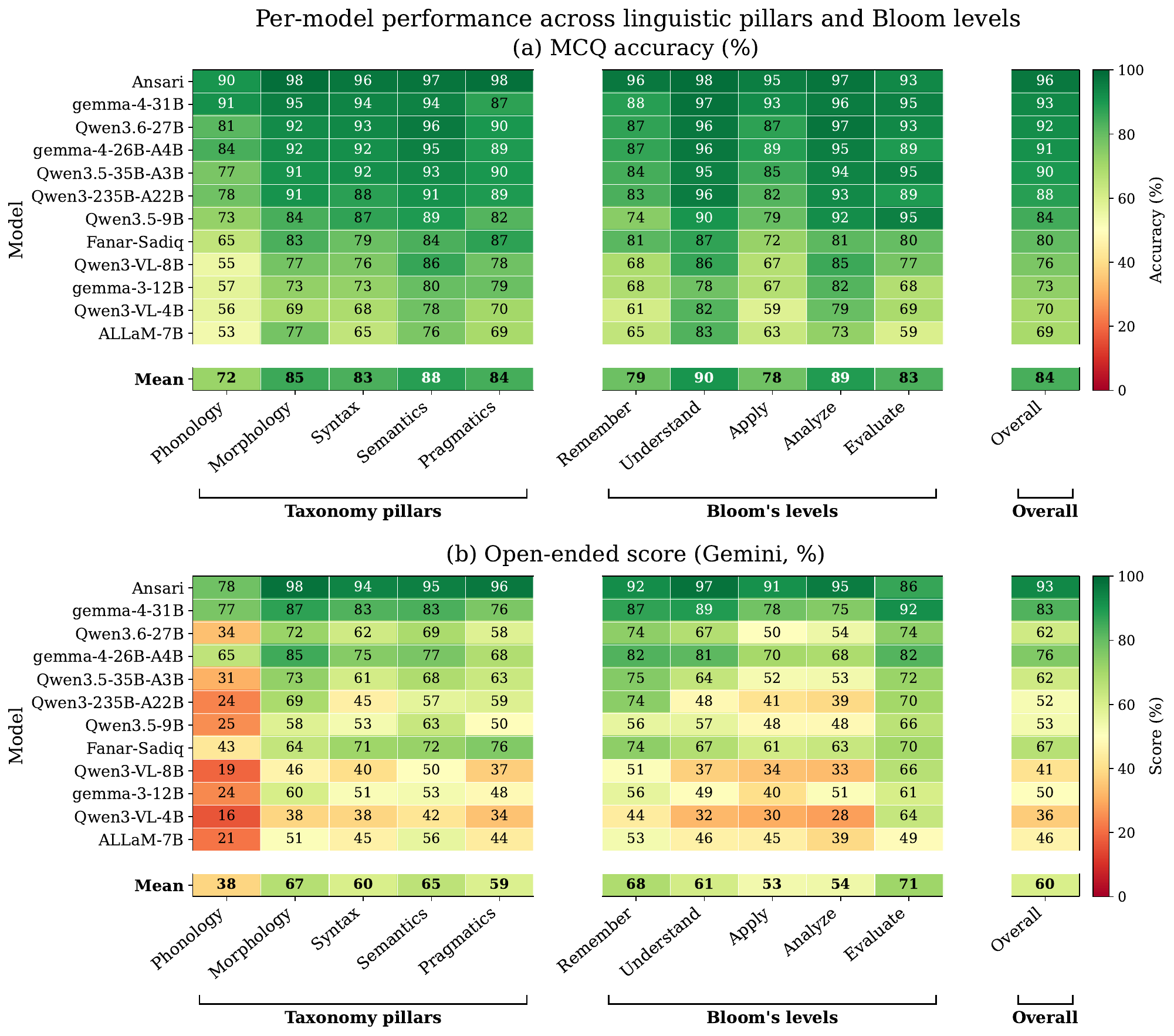}
  \caption{Per-model performance across the five linguistic pillars and the
  five Bloom levels, for (a) MCQ accuracy and (b) open-ended Gemini score.
  Models are ordered by overall MCQ accuracy. Phonology is the weakest
  pillar for nearly every system, especially in the open-ended setting.}
  \label{fig:results-heatmap}
\end{figure*}

 We evaluate the two formats with complementary metrics, all broken down
 by model and by Bloom level, with full breakdowns in
 Figure~\ref{fig:results-heatmap}. For
 \textbf{open-ended} answers, we use Gemini as an automated grader that
 scores each response from 0 to 5 given the question, a reference answer,
 and the model's response (Appendix~\ref{app:judge},
 p.~\pageref{app:judge}); we report this rating as a percentage of the
 5-point maximum, together with the
 proportion of responses scoring 80\% or higher. For \textbf{MCQ}, we report
 accuracy, the share of questions answered correctly, presented by Bloom
 level, by taxonomy content type, and by dataset; with four options,
 random guessing yields roughly 25\%. Finally, to test whether the two
 formats measure the same underlying ability, we compare model rankings
 under each using Kendall's $\tau$, where a strong correlation would
 indicate that both formats probe similar skills.

\section{Results}
\label{sec:results}

\begin{table}[t]
  \centering
  \small
  \begin{tabular}{lcc}
    \toprule
    \textbf{Model} & \textbf{MCQ Acc.\ (\%)} & \textbf{Open (\%)} \\
     & & {\scriptsize mean $\pm$ SD} \\
    \midrule
    Ansari            & \textbf{96.4} & \textbf{93.2}\,{\scriptsize$\pm$21.8} \\
    gemma-4-31B       & 93.3 & 82.5\,{\scriptsize$\pm$28.8} \\
    Qwen3.6-27B       & 91.8 & 62.1\,{\scriptsize$\pm$37.0} \\
    gemma-4-26B-A4B   & 91.2 & 75.7\,{\scriptsize$\pm$32.8} \\
    Qwen3.5-35B-A3B   & 89.7 & 62.1\,{\scriptsize$\pm$35.8} \\
    Qwen3-235B-A22B   & 88.3 & 52.3\,{\scriptsize$\pm$39.0} \\
    Qwen3.5-9B        & 84.5 & 53.1\,{\scriptsize$\pm$36.6} \\
    Fanar-Sadiq       & 80.1 & 66.7\,{\scriptsize$\pm$37.0} \\
    Qwen3-VL-8B       & 76.3 & 41.1\,{\scriptsize$\pm$37.2} \\
    gemma-3-12B       & 73.2 & 50.2\,{\scriptsize$\pm$38.2} \\
    Qwen3-VL-4B       & 69.7 & 36.0\,{\scriptsize$\pm$36.4} \\
    ALLaM-7B          & 69.5 & 46.4\,{\scriptsize$\pm$39.8} \\
    \bottomrule
  \end{tabular}
  \caption{Overall benchmarking results: multiple-choice accuracy and open-ended
  Gemini score, both as percentages, reported as mean $\pm$ standard deviation across items and
  sorted by MCQ accuracy. Per-pillar and per-Bloom
  breakdowns are in Figure~\ref{fig:results-heatmap}.}
  \label{tab:results}
\end{table}

On the \textbf{open-ended} evaluation, Gemini judge scores spanned 93.2\%
for Ansari down to 36.0\% for Qwen3-VL-4B (Table~\ref{tab:results}).
Ansari also scored 80\% or higher on 91.1\% of
items, ahead of the next system, gemma-4-31B, which averaged 82.5\% and
cleared that threshold on 75.6\% of items. Among the remaining systems, Fanar-Sadiq placed fourth (66.7\%),
and the largest model evaluated, Qwen3-235B-A22B, sat in the lower half
(52.3\%). The same items in \textbf{multiple-choice} form produced higher
scores for every system: Ansari again placed first (96.4\%), followed by
gemma-4-31B (93.3\%), while ALLaM-7B was the weakest at 69.5\%, still well
above the 25\% chance level, and the seven highest-scoring systems each
cleared 84\%.

The two orderings were closely aligned (Kendall's $\tau = +0.7273$; 57 of
66 model pairs concordant). Ansari and gemma-4-31B occupied the top two
positions in both, while Fanar-Sadiq moved up four places under the
open-ended setting (eighth in MCQ, fourth in open-ended). The full
per-pillar and per-Bloom breakdown is shown in
Figure~\ref{fig:results-heatmap}. Across pillars,
Phonology yielded the weakest scores for almost every system in both
settings, dropping below 34\% for most systems on the open-ended
questions, whereas Morphology and Semantics were consistently the
strongest. Across Bloom levels, scores peaked at the Understand level and
fell to their lowest at Apply and Analyze in both settings.
\section{Discussion and Conclusion}
Ansari, a system specialized for Islamic and Quranic content, ranked
first in both formats, and this domain specialization is reflected
directly in our results. The strong rank correlation between formats (Kendall's $\tau = +0.7273$)
indicates that, at the level of ordering systems, multiple-choice accuracy
and open-ended quality largely track the same underlying ability, which
might suggest that MCQ benchmarks could serve as an efficient proxy, though
confirming this would require a dedicated error analysis of the
LLM-as-judge scores. The
agreement is not perfect, however: Fanar-Sadiq rose four places under
open-ended grading not because it improved in absolute terms but because
other systems degraded more sharply when the answer choices were removed,
a gap that pure MCQ evaluation would hide.

The Bloom-level pattern mirrors a familiar profile of model cognition:
performance was highest at the Understand level and lowest at Apply and
Analyze, suggesting that current systems recognize and explain Quranic
phenomena more readily than they apply rules to new cases. Overall,
QuranicMMLU provides a fine-grained, linguistically grounded view of model
ability on Quranic Arabic and shows that format and cognitive demand both
shape measured performance. Future work includes broadening leaf coverage,
adding human scoring of open-ended answers, and evaluating further
Islamic-specialized systems.

\section{Limitations}
Our study has several limitations. (i)~The taxonomy is designed to be broad
and to cover a variety of phenomena within each pillar, but it does not
capture every linguistic rule or concept in those pillars. (ii)~Open-ended
responses are scored solely by an LLM judge without human scoring, which
may introduce model bias, including a possible preference for answers
resembling a judge's own model family. (iii)~We evaluate only open-weight
and open-access systems; closed-source commercial models were not assessed
because we did not have access to their APIs. (iv)~Our benchmark does not
yet include every relevant Islamic-specialized system, and some top-ranked
models on Islamic-domain leaderboards remain to be evaluated. (v)~Questions are generated by a single model (Claude Opus 4.7), so
despite human review some generation-model artifacts may remain.

\section{Ethical Statement}
We note the following ethical considerations. (i)~Our human reviewers were
not religious scholars; they reviewed questions using the grounding
datasets, reference books, online research, and their own judgment. This
dataset is intended solely for benchmarking model capabilities and should
not be treated as an authoritative source of religious fact. (ii)~We omit the Create level
of Bloom's taxonomy because generating new Quranic text raises clear
ethical concerns.


\bibliography{custom}

\appendix

\section*{Appendices}
The appendices follow the order of the pipeline in
Section~\ref{sec:taxonomy}ff: the taxonomy
(\ref{app:taxonomy-fig}, p.~\pageref{app:taxonomy-fig}), question generation
(\ref{app:prompt-stack}, p.~\pageref{app:prompt-stack}), the quality rubric
(\ref{app:rubric}, p.~\pageref{app:rubric}) and review interface
(\ref{app:review}, p.~\pageref{app:review}) used for annotation, the
open-ended judge prompt (\ref{app:judge}, p.~\pageref{app:judge}), the
Bloom-level verifier guidelines (\ref{app:guidelines},
p.~\pageref{app:guidelines}), and representative questions per pillar
(\ref{app:examples}, p.~\pageref{app:examples}).

\section{QuranicMMLU Taxonomy}
\label{app:taxonomy-fig}
\begin{figure}[H]
  \centering
  \includegraphics[width=\columnwidth]{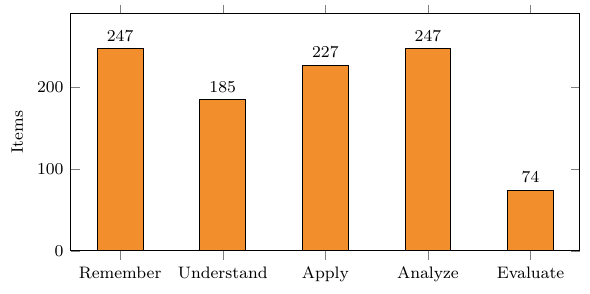}
  \caption{Overall Bloom-level distribution across all 980 items.}
  \label{fig:bloom-overall}
\end{figure}

\begin{figure*}[t]
  \centering
  \includegraphics[width=0.85\textwidth]{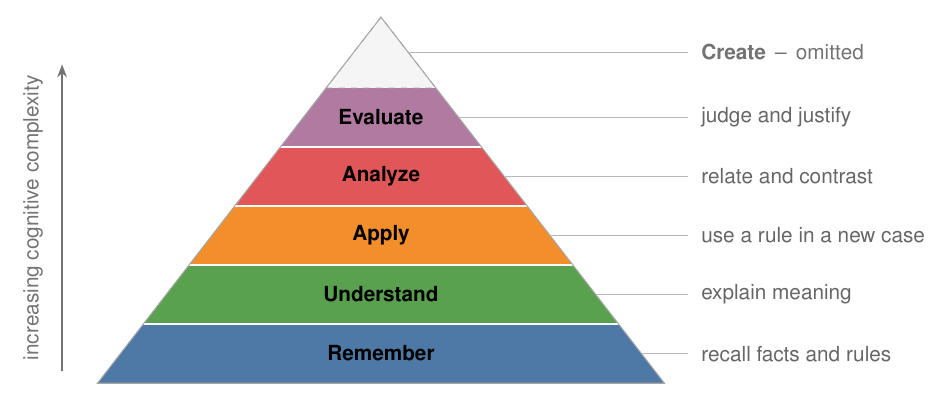}
  \caption{We stratify benchmark questions across five cognitive levels from Bloom's revised taxonomy (Remember through Evaluate), excluding the Create level, as synthesizing novel Quranic content would violate theological and scholarly principles in Islamic tradition.}
  \label{fig:bloom}
\end{figure*}

Figure~\ref{fig:bloom} shows the five Bloom's cognitive levels along which questions are stratified, and
Figure~\ref{fig:bloom-overall} the overall Bloom-level composition of the dataset.

\section{Question-Generation Prompt Pipeline}
\label{app:prompt-stack}
\begin{figure}[H]
  \centering
  \includegraphics[width=0.85\columnwidth]{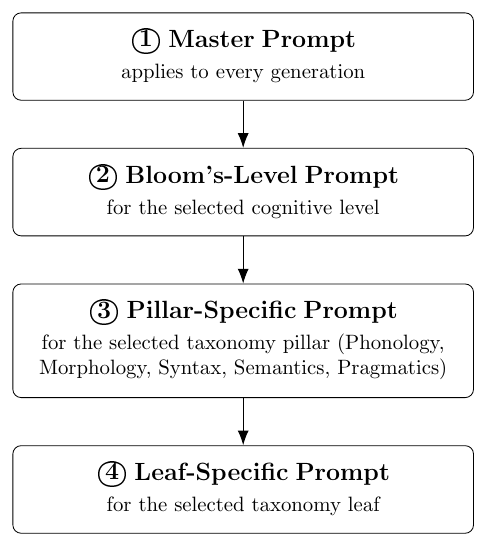}
  \caption{Hierarchical prompt-composition pipeline used to generate each
  benchmark question.}
  \label{fig:prompt-stack}
\end{figure}

Figure~\ref{fig:prompt-stack} shows the hierarchical prompt stack used to
compose each benchmark question from the taxonomy leaf, Bloom level, and
verse context.

\section{Question-Quality Rubric}
\label{app:rubric}
Every generated question is rated $1$--$5$ on four criteria to guide human
review (Figure~\ref{fig:quality-rubric}).

\begin{figure*}[t]
\begin{promptbox}[breakable=false]{Question-quality rubric used during annotation and review}
\small
\textbf{Specificity.} Does the question point to exactly one
defensible answer? $5=$ fully constrained (target word, recitation, and
referent fixed or inferable); $1=$ many defensible answers, none privileged
by the question.

\textbf{Answerability.} Can the question be answered from the
question, verse, translation, and leaf alone? $5=$ fully self-contained;
$1=$ depends on an artefact the model never sees, which marks the question
as unanswerable from the provided context.

\textbf{Leaf relevance.} Does the question genuinely test its
assigned leaf? $5=$ squarely on the leaf; $1=$ belongs to a different leaf.

\textbf{Clarity.} Is the wording unambiguous, so that two competent
readers interpret it identically? $5=$ a single interpretation; $1=$
incoherent wording.
\end{promptbox}
\caption{Question-quality rubric used to guide human review.}
\label{fig:quality-rubric}
\end{figure*}

\section{Human Review Interface}
\label{app:review}
Figure~\ref{fig:review} shows the custom web interface used for human review
and adjudication. For each item, reviewers see the question with the two LLM
judges' answers compared against the generator's reference
(Figure~\ref{fig:review-annotator}), the grounding evidence drawn from the
source datasets (Figure~\ref{fig:review-groundtruth}), and an adjudication
panel for recording the final decision (Figure~\ref{fig:review-adjudication}).

\begin{figure*}[p]
  \centering
  \begin{subfigure}{0.85\textwidth}
    \centering
    \includegraphics[width=\textwidth]{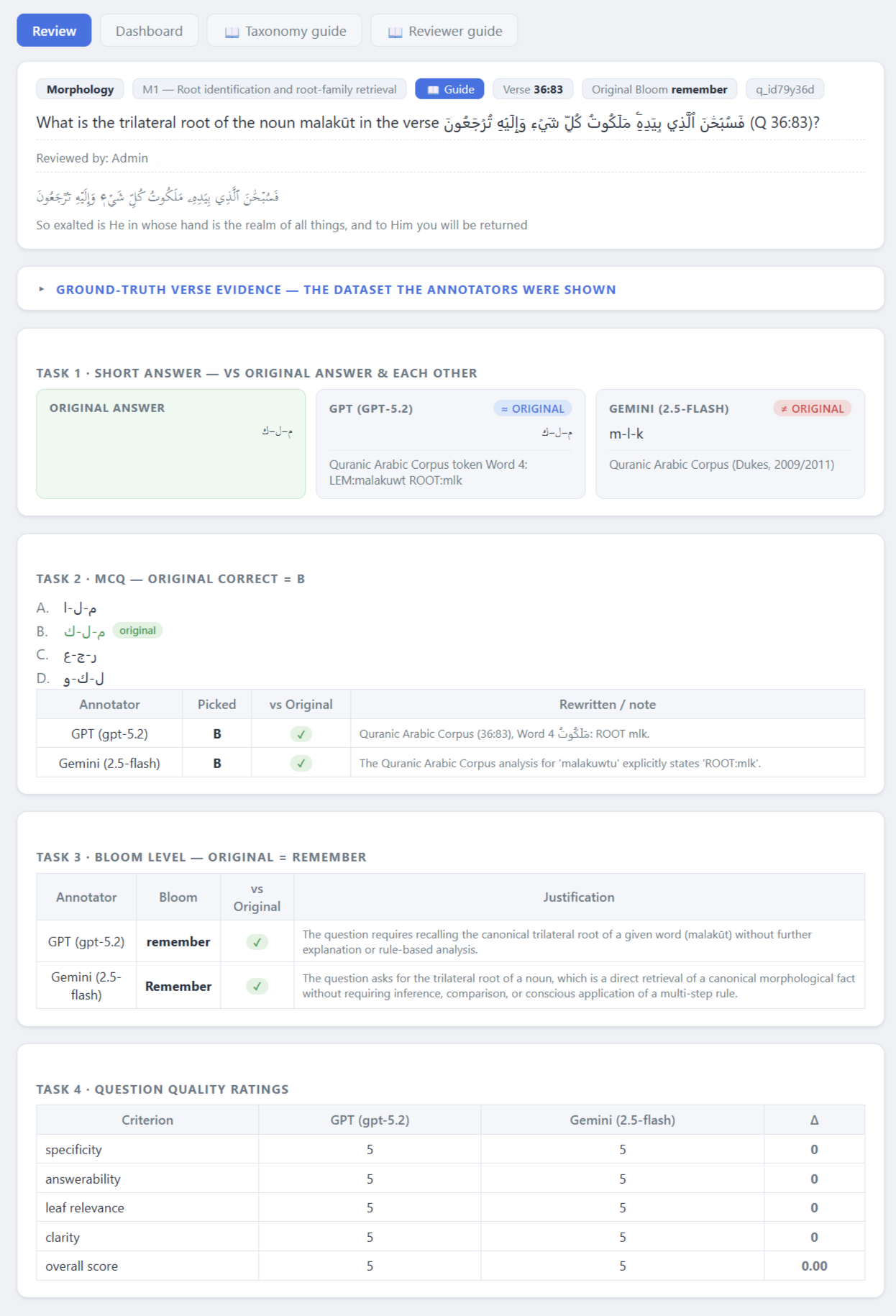}
    \caption{Annotation view: the question with its four tasks (short answer,
    MCQ, Bloom level, and quality ratings), comparing each LLM judge against
    the reference answer.}
    \label{fig:review-annotator}
  \end{subfigure}
  \caption{The custom web interface used for human review and adjudication.}
  \label{fig:review}
\end{figure*}

\begin{figure*}[p]\ContinuedFloat
  \centering
  \begin{subfigure}{0.85\textwidth}
    \centering
    \includegraphics[width=\textwidth]{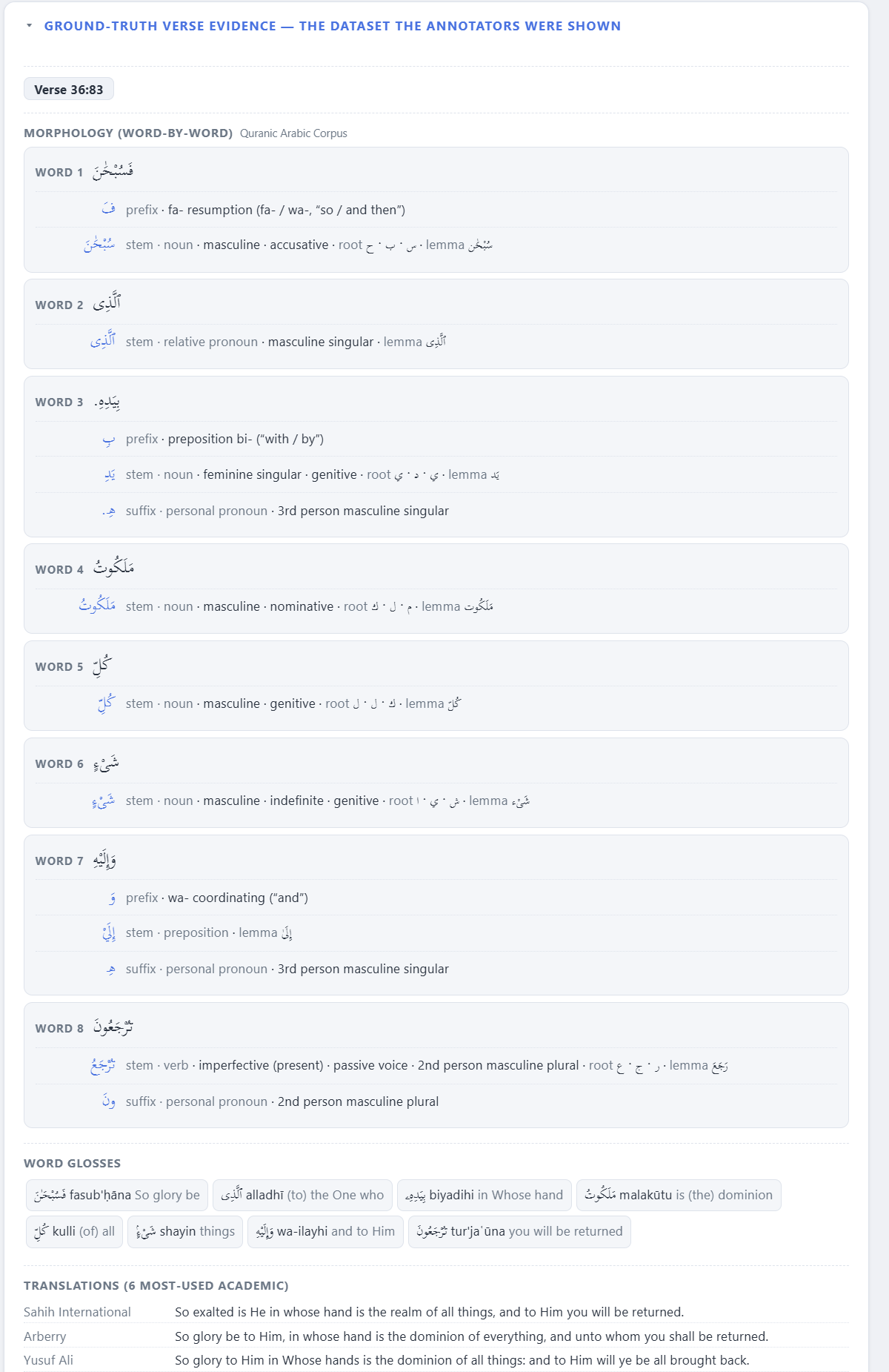}
    \caption{Grounding view: the word-by-word morphological evidence and
    translations shown to reviewers as ground truth for each verse.}
    \label{fig:review-groundtruth}
  \end{subfigure}
  \caption{(continued) Human review and adjudication interface.}
\end{figure*}

\begin{figure*}[p]\ContinuedFloat
  \centering
  \begin{subfigure}{0.8\textwidth}
    \centering
    \includegraphics[width=\textwidth]{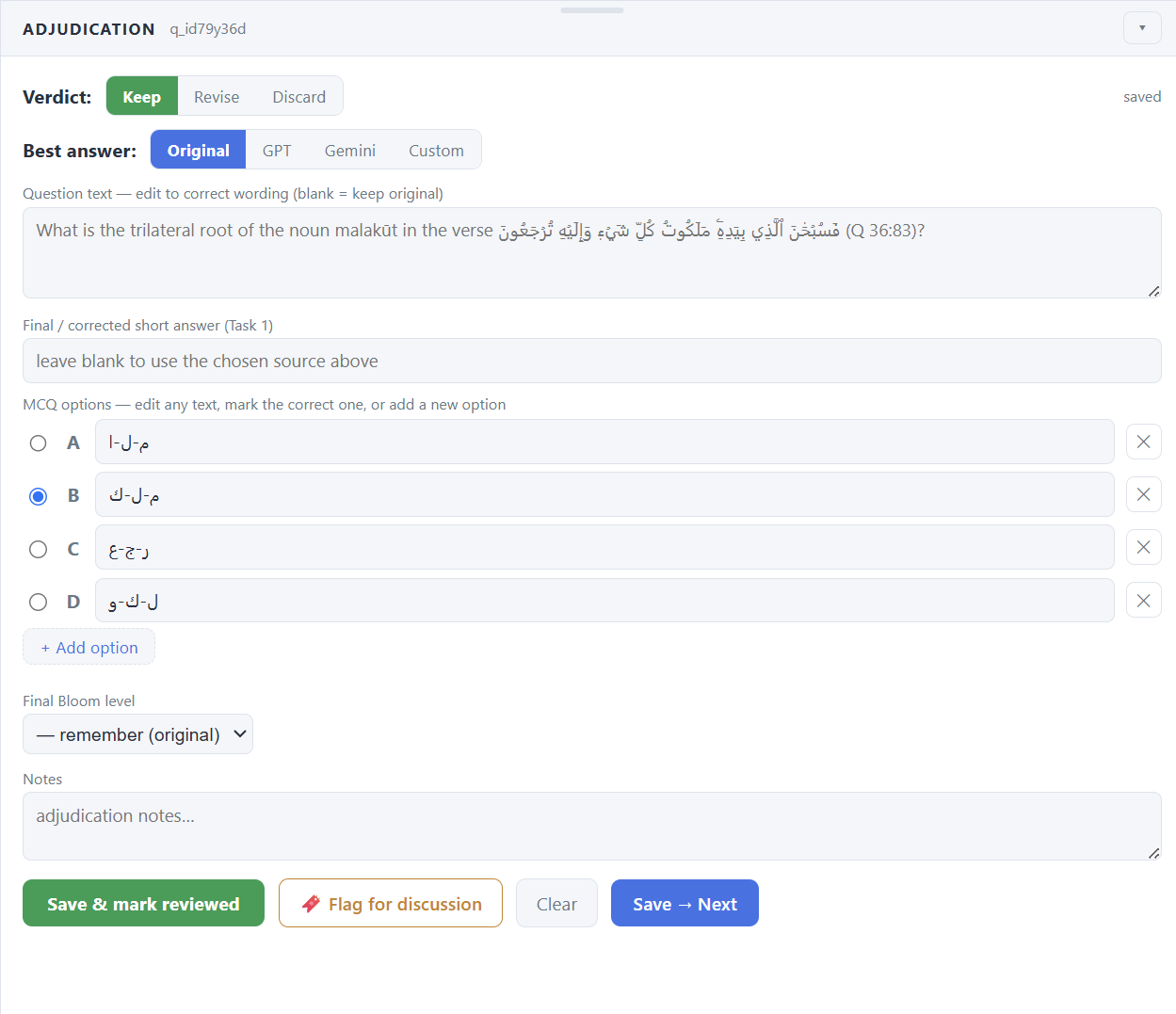}
    \caption{Adjudication panel: reviewers record a verdict (keep, revise, or
    discard), select or edit the final answer and options, set the Bloom
    level, and leave notes.}
    \label{fig:review-adjudication}
  \end{subfigure}
  \caption{(continued) Human review and adjudication interface.}
\end{figure*}

\section{LLM-Judge Scoring Prompt}
\label{app:judge}
Open-ended answers are scored automatically by an LLM judge against the
reference answer, using the prompt in Figure~\ref{fig:judge-prompt}.

\begin{figure*}[t]
\begin{promptbox}[breakable=false]{LLM-judge scoring prompt for open-ended answers}
\small
\textbf{Task.} Score each model's free-form answer against the reference
answer for an Arabic Quranic question.

\textbf{Scoring rubric ($0$--$5$), applied strictly.}
\textbf{5 -- Fully correct:} same meaning as the reference, no meaningful
omission; paraphrase and synonyms are fine.
\textbf{4 -- Mostly correct:} core answer right, one minor detail missing or
imprecise; would not mislead.
\textbf{3 -- Partially correct:} at least one correct key fact, but omits or
contradicts another.
\textbf{2 -- Mostly wrong:} predominantly incorrect, but contains at least
one relevant element.
\textbf{1 -- Barely relevant:} almost entirely wrong, but addresses the
right topic or domain.
\textbf{0 -- Wrong / empty:} factually wrong, hallucinated, off-topic,
refused, or empty.

\textbf{Label mapping.} correct $=$ 4--5; partial $=$ 2--3;
incorrect $=$ 0--1.

\textbf{Judging notes.} Judge meaning, not surface wording; Arabic synonyms
and paraphrases are acceptable. For a short entity or term, a near-exact
match scores $5$ and the correct meaning in other words scores $4$. Do not
penalise markdown formatting. Correct but irrelevant extra information is not
penalised unless it contradicts the reference.

\textbf{Inputs.} the question, the reference answer, and a block of model
answers.

\textbf{Output.} a single JSON object (no markdown fences) mapping each
model key to a score, label, and a one-sentence reason citing specific
content:\\[2pt]
{\footnotesize\ttfamily \{"model\_key": \{"score": 0-5, "label":
"correct|partial|incorrect", "reason": "..."\}\}}
\end{promptbox}
\caption{LLM-judge scoring prompt used to grade open-ended answers.}
\label{fig:judge-prompt}
\end{figure*}

\section{Verifier Guidelines for Bloom Levels}
\label{app:guidelines}
Reviewers assign each Bloom level by the cognitive operation the question
demands of the solver, not by how advanced its topic sounds.

\begin{promptbox}[breakable=false]{Verifier guidelines for assigning Bloom levels}
\begin{description}
  \setlength{\itemsep}{3pt}\setlength{\parsep}{0pt}
  \setlength{\leftmargin}{0pt}\setlength{\labelsep}{4pt}
  \item[Remember:] Recall an explicit fact from the verse or a standard
  reference, such as a root or a gloss.
  \item[Understand:] Explain or paraphrase a meaning or grammatical function
  in one's own words.
  \item[Apply:] Use a known rule or procedure on a specific instance, such
  as applying a tajw\={\i}d or grammar rule to a given word.
  \item[Analyze:] Break the verse down and derive a relationship or
  structure that is not stated, over more than one step.
  \item[Evaluate:] Weigh competing valid readings and justify which is
  stronger.
\end{description}
\end{promptbox}

The Bloom level was the most frequently corrected field during human review
(235 items; Section~\ref{sec:generation}), so the generated labels were
verified and corrected rather than accepted as produced.

\section{Representative Questions by Pillar}
\label{app:examples}
One representative question per pillar, with its human-verified Bloom level.
Arabic is transliterated; each expected answer follows the arrow.

\begin{promptbox}[breakable=false]{One question per pillar with its verified Bloom level}
\small
\begin{description}
  \setlength{\itemsep}{4pt}\setlength{\parsep}{0pt}
  \setlength{\leftmargin}{0pt}\setlength{\labelsep}{4pt}
  \item[Morphology (Remember):] What is the trilateral root of the verb
  ja\textquotesingle ala? $\rightarrow$ j-\textquotesingle-l.
  \item[Semantics (Understand):] Explain the difference in meaning between
  al-ra\d{h}m\=an and al-ra\d{h}\={\i}m in the basmala. $\rightarrow$ Both
  derive from the root r-\d{h}-m (mercy); al-ra\d{h}m\=an is the intensive
  form denoting all-encompassing mercy to every creature, while
  al-ra\d{h}\={\i}m denotes sustained, particular mercy.
  \item[Phonology (Apply):] Apply the n\=un s\=akinah rules to hudan min
  rabbihim (Q 2:5): what ruling governs the n\=un of min before rabbihim,
  and why? $\rightarrow$ Idgh\=am (assimilation) into the
  r\=a\textquotesingle{} with no ghunnah, since r\=a\textquotesingle{} is an
  idgh\=am letter that carries no nasalization.
  \item[Syntax (Analyze):] In iyy\=aka na\textquotesingle budu (Q 1:5),
  identify the role of iyy\=aka and explain what its position contributes.
  $\rightarrow$ It is a fronted direct object
  (maf\textquotesingle \=ul bihi muqaddam); placing the object before the
  verb restricts worship to God alone.
  \item[Pragmatics (Evaluate):] In Q 3:7, assess the two accepted stops,
  pausing at ill\=a All\=ah versus continuing to
  wa-l-r\=asikh\=una f\={\i} l-\textquotesingle ilm, and what each implies.
  $\rightarrow$ The first reading restricts knowledge of the
  mutash\=abih to God, the second extends it to those firmly grounded in
  knowledge; both are grammatically valid and attested.
\end{description}
\end{promptbox}

\end{document}